\documentclass[conference]{IEEEtran}
\IEEEoverridecommandlockouts
\usepackage{cite}
\usepackage{amsmath,amssymb,amsfonts}
\usepackage{algorithm}
\usepackage{algorithmic}
\usepackage{graphicx}
\usepackage{svg}
\usepackage{textcomp}
\usepackage{diagbox}
\usepackage{xcolor}
\usepackage{array}

\def\BibTeX{{\rm B\kern-.05em{\sc i\kern-.025em b}\kern-.08em
    T\kern-.1667em\lower.7ex\hbox{E}\kern-.125emX}}
\begin{document}

\title{Automated Generation of Precedence Graphs in Digital Value Chains for Automotive Production}

\author{\IEEEauthorblockN{1\textsuperscript{st} Cornelius Hake}
\IEEEauthorblockA{\textit{Dr. Ing. h.c. F. Porsche AG} \\
\textit{Hochschule Karlsruhe - University of Applied Sciences}\\
Stuttgart, Germany\\
cornelius.hake1@porsche.de}
\and
\IEEEauthorblockN{2\textsuperscript{nd} Christian Friedrich}
\IEEEauthorblockA{\textit{Hochschule Karlsruhe - University of Applied Sciences} \\
Karlsruhe, Germany \\
christian.friedrich@h-ka.de}
}

\maketitle

\begin{abstract}\\
This study examines the digital value chain in automotive manufacturing, focusing on the identification, software flashing, customization, and commissioning of electronic control units in vehicle networks. A novel precedence graph design is proposed to optimize this process chain using an automated scheduling algorithm, which combines structured data extraction from heterogeneous sources via natural language processing and classification techniques with mixed integer linear programming for efficient graph generation. The results show significant improvements in key metrics. The algorithm reduces the number of production stations equipped with expensive hardware and software to execute digital value chain processes, while also increasing capacity utilization through efficient scheduling and reduced idle time. Task parallelization is optimized, resulting in streamlined workflows and increased throughput. Compared to the traditional scheduling method, the automated approach has reduced preparation time by 50\% and reduced scheduling activities, as it now takes two minutes to create the precedence graph. The flexibility of the algorithm's constraints allows for vehicle-specific configurations while maintaining high responsiveness, eliminating backup stations and facilitating the integration of new topologies. Automated scheduling significantly outperforms manual methods in efficiency, functionality, and adaptability.
\medskip
\end{abstract}


\section{Introduction}
The development of connected vehicles is driven by the need to implement new customer functionalities and to meet the increasing demand for fully digitalized and networked vehicles. The required electrical/electronic (EE)-architectures are becoming increasingly complex, characterized by challenging interdependencies between electronic control units (ECUs) and their hierarchies \cite{01}. Modern vehicles contain more than 100 ECUs distributed across several bus technologies, including Controller Area Network (CAN), FlexRay, Local Interconnect Network (LIN), Low Voltage Differential Signaling (LVDS), and Media Oriented Systems Transport (MOST) bus types \cite{02}. This exponential growth in complexity necessitates automotive manufacturers to fundamentally transform their current processes, particularly with respect to the digital value chain (DVC) in automotive production \cite{03}.

Existing approaches to scheduling all DVC processes are reaching their limits due to two converging challenges: the unprecedented complexity of EE-architectures and the increasing diversification of vehicle derivatives. These challenges make manual production scheduling increasingly impractical and error-prone, requiring a comprehensive reassessment of the temporal and spatial process arrangements along the production line. Production planners are now faced with the task of scheduling thousands of interdependent processes, taking into account all relevant constraints and requirements for hundreds of potential vehicle configurations. Manually determining the optimal sequence of DVC processes for each derivative is no longer feasible. The current manual approach to scheduling DVC processes consumes significant engineering resources and introduces the potential for human error. An automated precedence graph allows this variability to be handled systematically by algorithmically determining optimal process sequences based on the individual configuration of each vehicle. By identifying the most efficient process sequences and opportunities for parallel processing, an automated precedence graph minimizes the number of production stations required.

\begin{figure}[b]
    \centerline{\includegraphics[width=3.3in]{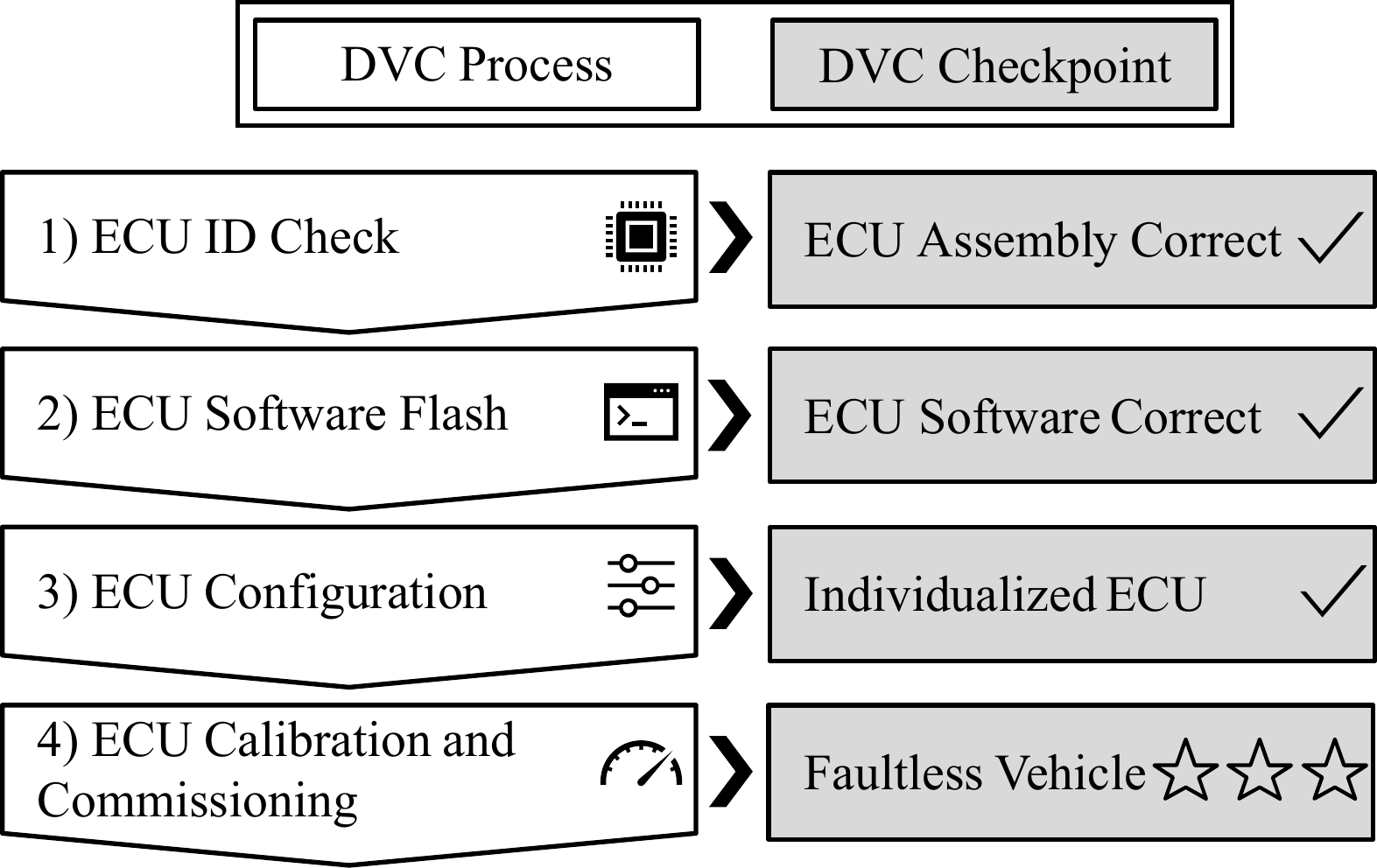}}
    \caption{Processes and checkpoints in the digital value chain (DVC) of automotive production}
    \label{fig:process}
\end{figure}

The DVC of a vehicle follows clearly defined processes. The general process is illustrated in Fig. \ref{fig:process}, and described below.
The ECU ID check (1) serves as the initial step in the DVC, focusing on the verification of various identifiers. Following the physical assembly of each ECU, this identification check confirms that the correct unit has been installed by comparing up to 20 different identifiers between actual and target values. The ID check identifies potential assembly problems and helps prevent incorrect parts from being assembled.
Afterwards, the ECU undergoes a software flashing process (2) where specific software is loaded onto the unit. Due to high flashing durations, only a limited number of ECUs are flashed during the assembly. The majority of ECUs are delivered with pre-installed software. Flashing ECUs during assembly offers the advantage for manufacturers to incorporate the most current software version directly into production.
The ECU then runs through a vehicle and customer-specific configuration (3). This configuration process adapts the ECU to meet the unique requirements and features specified by the customer.
Commissioning (4) involves several steps, such as calibrating sensors and actuators, recording measurements, performing functional tests, and reading and clearing fault memories. Beyond these operations, the final commissioning steps consist of software authenticity and compliance checks, as well as the creation of the assembly status documentation, which records all legally relevant information. The remainder of this paper is structured as follows. Section II reviews related work in DVC optimization and precedence graph generation. Section III details the methodology, including data preparation, information extraction, and the design of the scheduling algorithm. Section IV presents the evaluation results and discusses the improvements achieved. Finally, section V concludes the paper and outlines directions for future research.

\section{Related Work}
DVC in the automotive industry is highly regulated by patents. Patents exist for flashing ECUs \cite{patent1}, configuring ECUs \cite{patent2}, vehicle-specific configuring of ECUs \cite{patent3}, and commissioning of ECUs \cite{patent4}.

Create automated precedence graphs or learning graphs in automotive production is being researched in \cite{pred1} - \cite{pred3}. Precedence graphs are a methodology used to model complex interdependencies and temporal constraints in production processes \cite{pred4}. In the context of DVCs in automotive production, they provide a structured representation of sequential and parallelizable tasks. In \cite{pred1}, a learning approach for generating precedence graphs with a focus on assembly rather than DVC precedence graph generation is aimed at improving the efficiency and accuracy of assembly line scheduling in the automotive industry. The scope of \cite{pred2} is similar in investigating the optimization of assembly processes in the automotive industry by using multiple sources of information to create precedence graphs. Similar to the present study, \cite{pred2} uses information sources such as documented production plans containing sequences of tasks performed in the past, information about the modular structure of products and processes, and data from CAD systems containing geometric information about parts and their connections. A precedence graph generation technique based on system learning from past production sequences is described in \cite{pred3}. This technique builds a sufficient precedence graph to recognize near-optimal solutions even for a real segment of the automotive assembly line. A problem that \cite{pred3} describes is that not all constraint information is known. The reasons given are lack of documentation, decentralized or manual scheduling, and lack of digitalization, which are not present in this study.

Currently, there is no work on generating or automating precedence graphs for the DVC, which is due to the fact that most assembly lines only perform an end-of-line inspection, as only one derivative is handled in assembly. In this study, up to 28 derivatives from two series, divided into three drive variants, are processed on a single assembly line. With the increase in customer-specific individualization and the resulting highly individualized vehicles, this means a much greater potential for error, which is why the DVC differs fundamentally from that of other manufacturers. In this case, the DVC is highly integrated into the assembly process and takes place in parallel with manual activities on the vehicle.

In addition to the generation of precedence graphs, research is also being done on the scheduling of DVC processes.
The work of \cite{sched1} focuses on the development of an automated and flexible scheduling procedure for diagnostic tests in automotive manufacturing. The study proposes a mathematical model to handle the scheduling of diagnostic tests on parallel machines. The model considers multiple constraints such as time constraints, predecessor-successor relationships, mutual exclusion criteria, resource constraints, and status conditions. However, the study's flexible scheduling concept is only designed to handle different testing and commissioning tasks required for different vehicle configurations, and does not cover the entire DVC in automotive manufacturing.

Addressing the problem of maximizing the parallelization of DVC processes in a production station, \cite{sched2} proposes two scheduling algorithms for optimal parallelization of ECU software updates. The first is a Mixed-Integer Linear Program (MILP) and the second is a Hybrid Algorithm (HA), which combines a deterministic placement routine with a meta-heuristic approach. This reduces reprogramming time by up to 77\% compared to sequential updates. This shows the impact of parallelization on such problems.

Building on these insights, this paper investigates to what extent the automated generation of precedence graphs and their optimization via MILP-based scheduling can reduce planning effort, increase resource utilization, and support vehicle-specific configurations in DVC processes.

In particular, this work contributes to the following points:
\begin{enumerate}
    \item Automated generation of a DVC precedence graph to improve planning efficiency
    \item Benefits and challenges of using an algorithm to reduce scheduling times
    \item Proof of improved flexibility when using the algorithm to automate the DVC precedence graph
\end{enumerate}

\section{Methods and Materials}
\begin{figure}[t]
    \centerline{\includegraphics[width=3.3in]{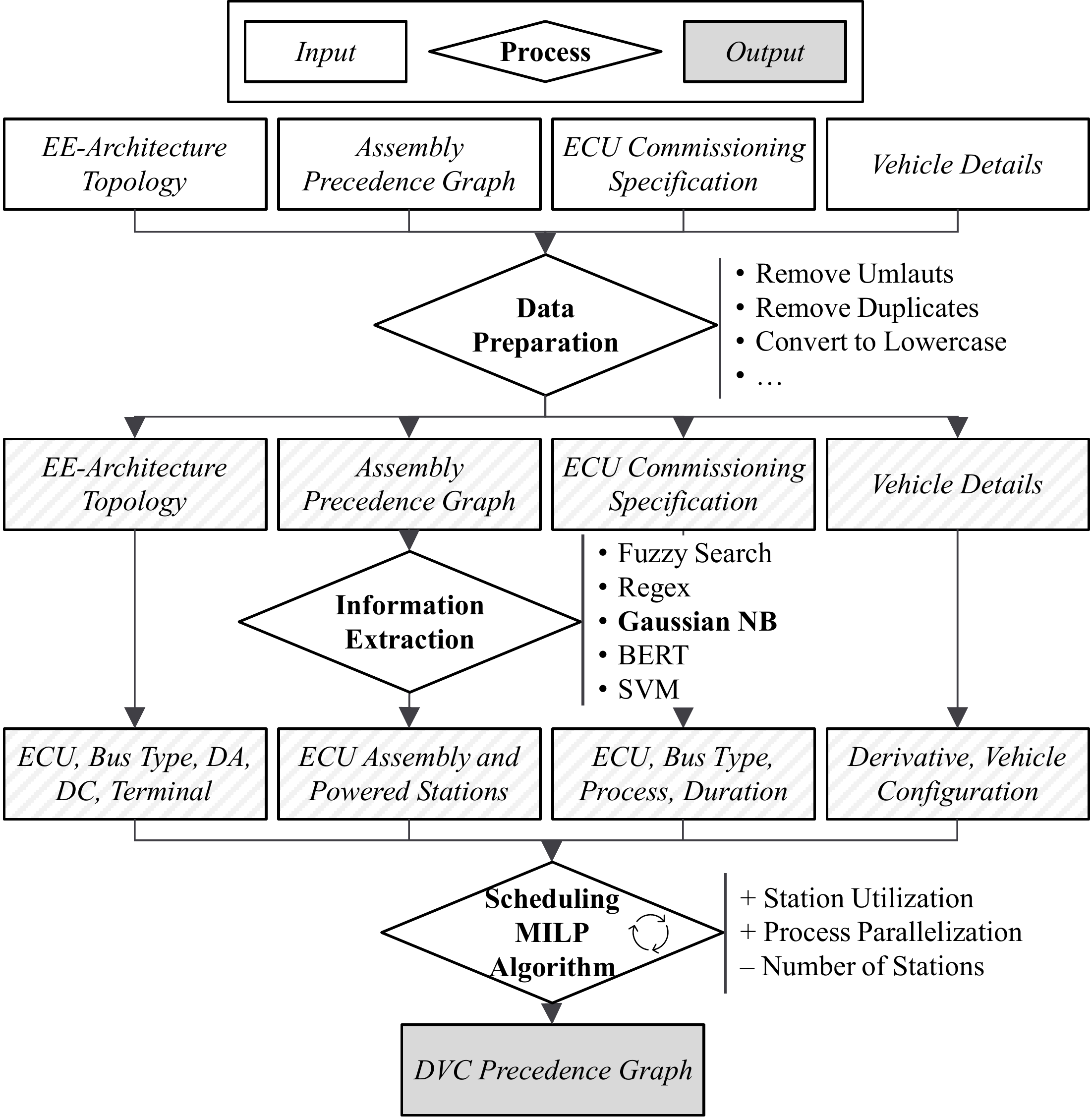}}
    \caption{Methodology from input (raw data) to output (precedence graph)}
    \label{fig:methodology}
\end{figure}

To automate the scheduling of the DVC processes, the first step is to analyze the existing data, including the data structure and its utilization within the DVC. Building on this, the second step presents the methods to achieve automation. The goal is to establish a consistent methodology, illustrated in Fig. \ref{fig:methodology}.

\subsection{Data for Precedence Graph Generation}
Essentially, the following information is required for a precedence graph generation: 
\begin{itemize}
    \item Number and type of ECUs, including bus type
    \item ECU sensitivity to terminal 15 or terminal 30
    \item ECU diagnostic class (DC) and diagnostic address (DA)
    \item ECU assembly station for each derivative
    \item Powered production stations 
    \item ECU commissioning steps and the required constraints and prerequisites
    \item Vehicle derivative and configuration
\end{itemize}
Terminal 15 and 30 are standardized according to DIN 72552 and define the electrical wiring in vehicles \cite{DIN72552}. Terminal 15 is the connection for the ignition, where the ECU receives power only when the ignition is activated. In contrast, terminal 30 is the connection for direct battery supply and indicates, that the ECU is constantly connected to the battery, regardless of whether the ignition is on or off. This is important information because it indicates whether an external power supply is required or whether the ignition is sufficient. The Diagnostic Class (DC) is an essential feature of each ECU that must be taken into account for scheduling. It defines further constraints for the DVC processes of each ECU in the complete system and is described in Tab. \ref{tab:diagnostic_classes}.

In the present study, the information is stored in four data sources. Tab. \ref{tab1} provides an overview of the data sources and the information they contain, along with an example. Information about the ECUs and their architecture is stored in the EE-architecture topology. The assembly precedence graph contains information about each vehicle configuration and represents the assembly sequence of the vehicle in a compact form as short text and in a detailed form as long text. The ECU engineering specifications contain information about ID checking, software flashing, and individual configuration, calibration, and final functionality checks for each ECU, including estimated durations and a detailed description of each process. The individual vehicle details such as configuration and derivative are stored in the production order.

\begin{table}[t]
\caption{Diagnostic Classes and their Descriptions}
\begin{center}
\resizebox{3.3in}{!}{
{\renewcommand{\arraystretch}{1}
\begin{tabular}{|c|l|}
\hline
\multicolumn{1}{|c|}{\textbf{Diagnostic Class}} & \multicolumn{1}{c|}{\textbf{Description}} \\ \hline
DC\_0 & Not diagnostic capable \\ \hline
DC\_1 & Diagnostic capable slaves \\ \hline
DC\_2 & Diagnostic and flashable slaves \\ \hline
DC\_3 & Standalone diagnostic and flashable ECUs \\ \hline
DC\_4 & Standalone diagnostic and flashable ECUs \\& with diagnostic capable slaves \\ \hline
\end{tabular}}}
\label{tab:diagnostic_classes}
\end{center}
\end{table}

\begin{table}[t]
\caption{Overview of Data Sources}
\begin{center}
\resizebox{3.3in}{!}{
{\renewcommand{\arraystretch}{1}
\begin{tabular}{|>{\raggedright\arraybackslash}p{1.1in}|>{\raggedright\arraybackslash}p{1.1in}|>{\raggedright\arraybackslash}p{1.1in}|}
\hline
\multicolumn{1}{|c|}{\textbf{Table}} & \multicolumn{1}{c|}{\textbf{Data}} & \multicolumn{1}{c|}{\textbf{Example}} \\ 
\hline
EE-architecture Topology & ECU, Bus Type, DA, DC, Terminal & Gateway, CAN, 0x70, 5, 15 \\
\hline
Assembly precedence Graph & PSL, Station, Short Text, Long Text & Basis Min, 4, Door Assembly, ... \\
\hline
ECU Commissioning Specification & ECU, Duration, Planned Station, Commissioning & Gateway, 20s, 12, ID Check\\
\hline
Vehicle Details & Product Key, Configuration Code & Basis Min, 1A1 - Interior Mirror black\\
\hline
\end{tabular}}}
\label{tab1}
\end{center}
\end{table}

\subsection{Data Preparation and Information Extraction}
Since the data sources containing the information are not standardized due to the actual manual creation, existing information extraction methods as a subcluster of natural language processing must be applied to further process the data, such as text cleaning, which removes umlauts, removes duplicates, and converts all text to lowercase \cite{ie1}. All steps described are based on Python code version 3.12 and run on a standard laptop with a 12th generation Intel\textregistered{} Core\texttrademark{} i7-1265U with 32GB RAM and no GPU support.

It is essential to extract from the assembly precedence graph both the stations where the ECUs are installed and the stations where the vehicles are externally powered. The information is present in both short and long text, but the wording and expression vary depending on the responsible assembly planner. Therefore, a generally applicable methodology is sought to ensure reliable information extraction.
To address this challenge, five distinct methodologies are investigated: two classical word matching approaches (Fuzzy search and Regular expressions) and three classification methods (BERT, Gaussian Naive Bayes, and Support Vector Machine) \cite{ie2}.
For the fuzzy matching approach, the FuzzyWuzzy Python package implementing Levenshtein distance is employed to match unique ECUs from the vehicle topology in the assembly precedence graph despite potential misspellings or alternative formulations \cite{fuzzy1}. This implementation requires three specific inputs: 
\begin{itemize}
    \item The assembly precedence graph as search target
    \item Search terms from a list of unique ECUs derived from the topology
    \item A matching threshold set at 90\% 
\end{itemize}
Successful matches are only returned when the searched ECU appears in the assembly precedence graph in combination with variations of 'contact' or 'install' exceeding the defined threshold \cite{fuzzy2}.
The regular expression methodology defines a match as the presence of one of the unique ECUs within a given row of the assembly precedence graph, in conjunction with the same words used for the fuzzy search.

In contrast to these classical methods, the three classification approaches follow a different methodology, classifying sentences from the assembly precedence graph into those where ECUs are assembled and those where they are not, to subsequently determine the corresponding station. This requires creating a labeled dataset of 1000 manually labeled rows from the assembly precedence graph, with 250 rows representing ECU assembly, 250 representing powered stations, and 500 representing neither condition.
For the BERT implementation, fine-tuning is performed on a pre-trained DistilBERT model with the prepared dataset. This process involves training the pre-trained DistilBERT model with the appropriate labels to optimize performance for the task \cite{bert1}, \cite{bert2}. The selection of DistilBERT is deliberate as it is more resource efficient than the original BERT model, allowing the system to make accurate predictions and adapt to the specific needs of text classification \cite{distilbert}.
Text classification with Gaussian Naive Bayes is implemented in conjunction with the term Frequency Inverse Document Frequency (TF-IDF)-vectorizer with training done in the same way as for the BERT model \cite{gaussiannb1}. Similarly, the Support Vector Machine classification also utilizes the TF-IDF-vectorizer, after which the SVM model is trained \cite{svm1}.

An experimental evaluation is conducted to determine the most appropriate method for identifying both ECU installation stations and powered stations in the assembly precedence graph across different vehicle derivatives. The results, shown in Tab. \ref{tab2} demonstrate that the Gaussian Naive Bayes algorithm can be reliably used to determine the stations where the ECUs are installed and the vehicle is powered.

\begin{table}[hb]
\caption{Results of the experiment to extract information from the assembly precedence graph}
\begin{center}
\resizebox{3.3in}{!}{
\begin{tabular}{|c|cc|ccc|}
\hline
\diagbox{\textbf{Process}}{\textbf{Methods}} & \multicolumn{2}{c|}{\textbf{Classic}}                       & \multicolumn{3}{c|}{\textbf{Classification}}                                           \\ \hline
                          & \multicolumn{1}{c|}{\textbf{Fuzzy Search}} & \textbf{Regex} & \multicolumn{1}{c|}{\textbf{DistilBERT}} & \multicolumn{1}{c|}{\textbf{Gaussian NB}} & \textbf{SVM}   \\ \hline
\textbf{ECU Assembly}     & \multicolumn{1}{c|}{94.32\%}                 & 2.56\%           & \multicolumn{1}{c|}{90.67\%}         & \multicolumn{1}{c|}{99.87\%}                & 94.34\% \\ \hline
\textbf{Powered Stations} & \multicolumn{1}{c|}{96.28\%}                 & 40.90\%          & \multicolumn{1}{c|}{92.36\%}         & \multicolumn{1}{c|}{97.07\%}                & 95.47\% \\ \hline
\end{tabular}}
\label{tab2}
\end{center}
\end{table}

\subsection{Prerequisites and Constraints in DVC}
The algorithm must take into account various prerequisites and constraints such as resource capacity, time limitations, and material availability to generate an optimal precedence graph, which will be discussed further.
\begin{itemize}
    \item A prerequisite for all bus types is an available and reliable power supply. 
    \item CAN bus and Flexray must be terminated at both ends with a 120 Ohm resistor to avoid reflections and must have a common time base for all nodes to ensure synchronization \cite{02}. 
    \item To start a FlexRay cluster, at least two nodes must send startup frames. The nodes that send the startup frames are called cold starter nodes. After receiving the first frame, the subsequent cold starter waits for the second frame. Using the information from the cycle and the slot, the subsequent cold starter synchronizes its clock with the leading cold starter. Once the nodes are synchronized, normal FlexRay communication begins \cite{flexray}. 
    \item The LIN bus does not require a termination resistor because it is typically designed as a master-slave system. A master device initiates communication with one or more slave devices. Communication on the bus can only take place if a master is connected \cite{lin}.
    \item The MOST bus must be terminated at both ends with a 75 Ohm resistor to avoid reflections, this is achieved by designing the MOST bus as a loop. Both adding and removing a MOST device is possible without any problems due to the plug-and-play functionality. Star or other topologies can be set up virtually \cite{most}.
    \item Due to the technological conditions, different bus technologies can be processed in parallel, taking into account requirements such as the presence of master and slave ECUs or fully connected MOST bus ring lines.
    \item The DC constraints from Tab. \ref{tab:diagnostic_classes} must be met.
\end{itemize}

In addition, each process has its own set of rules. These are considered in more detail, while configuration and calibration and commissioning are summarized, as the same rules apply to these processes.

\subsubsection{ECU ID Check}
For ECUs with a DC less than three, the higher-level ECU must be fully configured before performing an ID check. For ECUs with DC three or higher, the higher-level ECU only needs to be capable of performing ID checks. In both scenarios, the specific requirements of the BUS system must be considered, including installation of necessary components like Flexray cold starters. It is critical to avoid BUS interference when contacting an ECU on an active bus.

\subsubsection{ECU Software Flash}
When flashing ECUs with DC less than or equal to two, the higher-level ECU must be fully commissioned to its generic extent. For ECUs with DC three or higher, only an ID check of the master ECU is required. The flash process is sensitive to interruptions—power supply failures, terminal 15 changes, or BUS faults will abort the process. Therefore, no ECUs on the currently used BUS should be contacted during flashing.

\subsubsection{ECU Configuration, Calibration and Commissioning}
For the configuration process, higher-level ECUs in the topology must be appropriately prepared, either flashed or ID checked, depending on the DC. To maintain system stability, no ECU installations should occur on the same bus during configuration. Additionally, parallel processes on the master ECU or on slaves of the same master (for DC less than 3) are not permitted. Various signals including vehicle speed $v$, transmission parking $p$, and Vehicle Protective Environment $VPE$ signal are required for proper commissioning.

\subsection{Design of the Algorithm}
Now that all the necessary information, prerequisites, constraints, and the DVC processes are defined, the scheduling algorithm can be designed. The given problem is classified as a Mixed Integer Linear Programming (MILP) optimization problem, which is defined by a set of indices, parameters, decision variables, an objective function, and constraints \cite{milp}. The goal is to ensure parallel bus usage and full station utilization within the specified time limits by minimizing the total number of stations. All definitions are summarized in pseudocode Alg. \ref{alg:definitions} and the main code is shown in Alg. \ref{alg:main}.

The process begins with the collection of input data, which includes the vehicle configuration and the ECUs $e$ for that configuration. Then, the powered stations are identified, followed by the ECU assembly stations $s$. Next, the relevant bus types $b$ are identified, along with the DVC processes $p_{1,2,3}$ and their estimated duration $d$ for each ECU. The next step is to assign an ID check station for each ECU, taking into account the assembly station and the duration of the ID check. The software flash stations are then assigned, taking into account a correct ID check and the cumulative duration for ID check and software flash.

\begin{algorithm}[hb]
\caption{Scheduling Algorithm Definitions}
\label{alg:definitions}
\begin{algorithmic}[1]
\STATE \textbf{//Sets and Indices:}
\STATE $e \in E$: Set of all ECUs
\STATE $b \in B$: Set of all bus types
\STATE $s \in S$: Set of all stations
\STATE $p \in P$: Process types ($p_1$: ID check, $p_2$: software flash, $p_3$: configuration)
\STATE \textbf{//Parameters:}
\STATE $d_{e,b,p}$: Duration for process $p$ on ECU $e$ on bus $b$
\STATE $CT$: Cycle time limit
\STATE $PS_s$: Binary, 1 if station $s$ is powered
\STATE $AE_{e,s}$: Binary, 1 if ECU $e$ is assembled at station $s$
\STATE $DC_{e,b}$: Binary, 1 if DC constraints for ECU $e$ and bus $b$ are met
\STATE $V_{e,b}$: Binary, 1 if v signal is required for $e$ and $b$
\STATE $VPE_{e,b}$: Binary, 1 if VPE signal is required for $e$ and $b$
\STATE $P_{e,b}$: Binary, 1 if P signal is required for $e$ and $b$
\STATE $HV_{e,b}$: Binary, 1 if HV connection is required
\STATE \textbf{//Decision Variables:}
\STATE $x_{e,b,p,s}, y_{b,s}, z_s$: Binary decision variables
\STATE $CD_{b,s}$: Cumulative duration for bus $b$ at station $s$
\STATE $s_{e,b,p}$: Station ID assigned to process $p$ for $e$ on $b$
\STATE \textbf{//Objective Function:}
\STATE $f = \min (\alpha \cdot \sum_{s \in S} z_s + \beta \cdot \sum_{s \in S} \sum_{b \in B} CD_{b,s})$
\STATE \textbf{//Performance Metrics:}
\STATE Utilization: $U = \frac{1}{\sum_{s \in S} z_s} \sum_{s \in S} \sum_{b \in B} \frac{CD_{b,s}}{CT} \cdot y_{b,s}$
\STATE Parallelization: $P = \frac{1}{\sum_{s \in S} z_s} \sum_{s \in S} \frac{\sum_{b \in B} y_{b,s}}{|B|}$
\end{algorithmic}
\end{algorithm}

 Finally, the ECU configuration, calibration, and commissioning stations are assigned, taking into account a correct ID check and software flash, as well as the cumulative time for ID check, software flash, and configuration.
For each valid station, a check is performed to determine if the ECU duration combined with the cumulative duration is less than the cycle time $CT$ and if all constraints and prerequisites are met. If not, the station is incremented to the next valid station and the process is repeated.

\begin{algorithm}[t]
\caption{Main Scheduling Algorithm}
\label{alg:main}
\begin{algorithmic}[1]
\STATE \textbf{//Main Algorithm:}
\STATE Initialize $z_s = 0$ $\forall$ $s \in S$, $CD_{b,s} = 0$ $\forall$ $b \in B, s \in S$
\FOR{$s \in S$}
    \FOR{$b \in B$}
        \STATE $CD_{b,s} = 0$
        \FOR{$e \in E$}
            \FOR{$p \in P$}
            \IF{$CD_{b,s} + d_{e,b,p} \leq CT$ 
            \STATE{AND} $CheckConstraints(e,b,p,s)$ (cf. III. B.) returns true}
                \STATE $z_s = 1$
                \STATE $CD_{b,s} = CD_{b,s} + d_{e,b,p}$
            \ELSE
                \STATE $s \leftarrow s + 1$
            \ENDIF
            \ENDFOR
            \STATE Ensure $s_{e,b,p_1} \leq s_{e,b,p_2} \leq s_{e,b,p_3}$
        \ENDFOR
    \ENDFOR
\ENDFOR
\STATE $f = \alpha \cdot \sum_{s \in S} z_s + \beta \cdot \sum_{s \in S} \sum_{b \in B} CD_{b,s}$
\STATE \textbf{return} $f$
\end{algorithmic}
\end{algorithm}

\section{Results}
\begin{figure}[hb]
    \centerline{\includegraphics[width=3.1in]{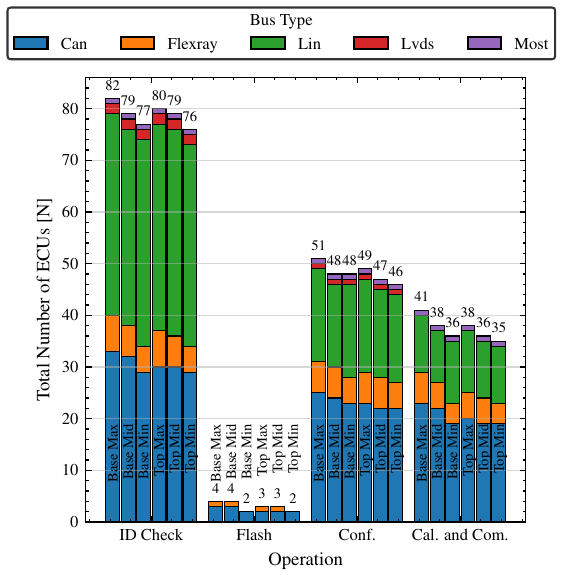}}
    \caption{Total Number of ECUs by DVC process and Bus Type}
    \label{fig:busdistribution}
\end{figure}

For evaluation, we compare the traditional manual scheduling and the developed automated scheduling algorithm. Six vehicles are examined, divided into two derivatives, with each derivative configured once with minimum, medium, and maximum configuration levels. The "Base" derivative represents a standard model, while the "Top" derivative represents the most valuable model. The "Base Min" configuration, represented by 175 features, corresponds to the minimum configuration variant that can be ordered by a customer, while "Base Mid" (181 features) represents the most frequently ordered combination in the base derivative. Conversely, "Base Max" (184 features) is the maximum configurable option within the Base derivative. The same is true for the "Top" derivative, where "Top Max" (189 features) represents the highest configuration level within the product portfolio, which is the most complex in assembly, while "Top Mid" (176 features) represents the most frequently ordered variant and "Top Min" (159 features) represents the absolute minimum variant. The number of features ordered by the customer has a direct impact on the complexity of the vehicle and the number of ECUs assembled, as well as the ECU assembly sequence and individual calibration and commissioning processes. In this context, Fig. \ref{fig:busdistribution} shows the total number of ECUs for each of the above mentioned derivatives in relation to the DVC processes, where conf. is the abbreviation for configuration and Cal. \& Com. is the abbreviation for calibration and commissioning. 

\begin{figure}[hb]
    \centerline{\includegraphics[width=3.1in]{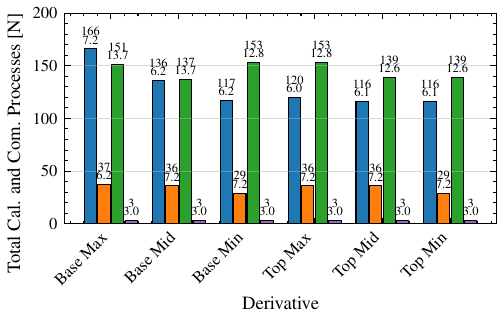}}
    \caption{Total Number of Calibration and Commissioning per ECU and Bus Type including average of processes per ECU}
    \label{fig:busoperationsforcommissioning}
\end{figure}

The largest number of ECUs requires the ID Check process, which reaches about 80 ECUs depending on the derivative. On the other hand, the Flash process is only for 4 ECUs. The configuration process shows 51 ECUs, while the calibration and commissioning processes vary between 35 and 41 ECUs. For ID check, flash and configuration, one process is performed for each ECU, while for calibration and commissioning, several processes are included in this process, which is further detailed in Fig. \ref{fig:busoperationsforcommissioning}. This is due to the individual calibration and commissioning steps for each control unit depending on the respective derivative and the functions ordered by the customer.


\begin{figure}[ht]
    \centerline{\includegraphics[width=3.3in]{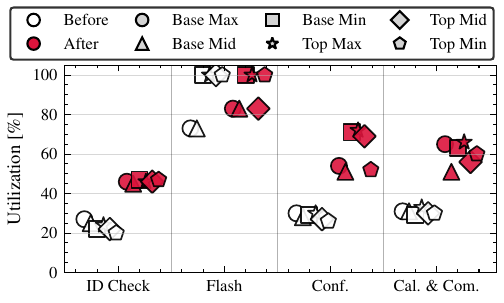}}
    \caption{Station Utilization by Process before and after the scheduling algorithm}
    \label{fig:utilization}
\end{figure}

\begin{figure}[ht]
    \centerline{\includegraphics[width=3.3in]{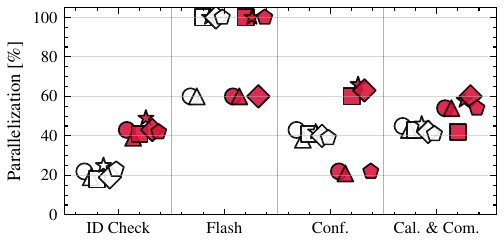}}
    \caption{Station Parallelization by Process before and after the scheduling algorithm}
    \label{fig:parallelization}
\end{figure}

The result for the required stations by derivative is shown in Tab. \ref{tab:stations}. Additionally, Fig. \ref{fig:utilization} and Fig. \ref{fig:parallelization} illustrate the results for utilization and parallelization, respectively.

\begin{table}[ht]
\centering
\caption{Stations before ($Sb$) and after the scheduling algorithm ($Sa$)}
\label{tab:stations}
\resizebox{3.3in}{!}{%
\begin{tabular}{|c|cc|cc|cc|cc|}
\hline
                    & \multicolumn{2}{c|}{\textbf{ID Check}}         & \multicolumn{2}{c|}{\textbf{Flash}}            & \multicolumn{2}{c|}{\textbf{Conf.}}            & \multicolumn{2}{c|}{\textbf{Cal. \& Com.}}    \\ \hline
\textbf{Derivative} & \multicolumn{1}{c|}{\textbf{$Sb$}} & \textbf{$Sa$} & \multicolumn{1}{c|}{\textbf{$Sb$}} & \textbf{$Sa$} & \multicolumn{1}{c|}{\textbf{$Sb$}} & \textbf{$Sa$} & \multicolumn{1}{c|}{\textbf{$Sb$}} & \textbf{$Sa$} \\ \hline
\textbf{Base Max}   & \multicolumn{1}{c|}{9}           & 7           & \multicolumn{1}{c|}{3}           & 3           & \multicolumn{1}{c|}{8}           & 10          & \multicolumn{1}{c|}{22}          & 13          \\ \hline
\textbf{Base Mid}   & \multicolumn{1}{c|}{9}           & 7           & \multicolumn{1}{c|}{3}           & 3           & \multicolumn{1}{c|}{8}           & 10          & \multicolumn{1}{c|}{21}          & 13          \\ \hline
\textbf{Base Min}   & \multicolumn{1}{c|}{9}           & 7           & \multicolumn{1}{c|}{2}           & 2           & \multicolumn{1}{c|}{8}           & 6           & \multicolumn{1}{c|}{21}          & 12          \\ \hline
\textbf{Top Max}    & \multicolumn{1}{c|}{9}           & 7           & \multicolumn{1}{c|}{2}           & 2           & \multicolumn{1}{c|}{8}           & 6           & \multicolumn{1}{c|}{21}          & 12          \\ \hline
\textbf{Top Mid}    & \multicolumn{1}{c|}{9}           & 7           & \multicolumn{1}{c|}{2}           & 3           & \multicolumn{1}{c|}{8}           & 6           & \multicolumn{1}{c|}{21}          & 12          \\ \hline
\textbf{Top Min}    & \multicolumn{1}{c|}{9}           & 7           & \multicolumn{1}{c|}{2}           & 2           & \multicolumn{1}{c|}{8}           & 10          & \multicolumn{1}{c|}{21}          & 13          \\ \hline
\end{tabular}%
}
\end{table}

The ID check process serves as the first verification step in the DVC process. The process is improved in both parallelization and utilization parameters. The parallelization increases from about 20\% to 40-45\%, while the utilization increases from 25\% to 45-50\% after the scheduling algorithm. At the same time, the number of required stations decreases uniformly from 9 to 7 across all derivatives, indicating improved resource efficiency that translates into monetary benefits. Base and Top derivatives show comparable improvement patterns in this process.

For the software flash process, the scheduling algorithm does not provide parallelization improvements, but some utilization gains for derivatives that did not initially reach 100\%, from 75\% to 80-85\%. Station requirements remain stable at 2-3 stations depending on the derivative type. Base derivatives maintain 3 stations after implementation, while Top Max and Top Mid derivatives require 2-3 stations and Base Min and Top Min consistently utilize 2 stations.

The configuration process faces some of the most significant changes. The number of stations varies by derivative: Base Max and Base Mid derivatives increase from 8 to 10 stations, while Top derivatives decrease from 8 to 6 stations, with the exception of Top Min, which increases to 10 stations. The parallelization metrics improve from 40\% to 60-65\%, with utilization increasing from about 30\% to 50-70\%, demonstrating effective use of the additional stations. The lower parallelization values for derivatives with increased station counts are explained by cycle time constraints. Despite an overall reduction in the number of stations required for all DVC processes, the configuration processes in particular required more stations. The algorithm allowed for a better distribution of tasks across stations, resulting in a more balanced workload. The varying, but significantly increased, parallelization rates indicate that the algorithm was successful in allowing more tasks to be processed simultaneously.

In the calibration and commissioning processes, there is a significant reduction in the number of stations, from 21-22 to 12-13 stations. This reduction indicates a major reorganization. As a result, parallelization improves from 40-45\% to 55-60\%, and utilization increases from 30-35\% to 55-65\%. The algorithm's ability to streamline processes and reduce redundancy likely contributed to this increase in efficiency. The parallelization rate also showed improvement, with a significant increase in the ability to perform tasks concurrently.

\section{Conclusion and Outlook}
The study demonstrates that the automated scheduling algorithm successfully achieves its key objectives of reducing station count, optimizing utilization, and improving process parallelization. Compared to the traditional manual scheduling approach requiring two weeks of preparation, the algorithm reduces this to five days, with precedence graph creation accelerated from ten days to just two minutes. The algorithm eliminates the need for continuous adjustments and back-up stations while maintaining high responsiveness and efficiency across all DVC processes.

Future research should be derived from the identified limitations and focus on enhancing parallelization capabilities in processes where optimization remains constrained. The input data used to generate the precedence graph is derived from heterogeneous and partially unstructured sources, which may affect the robustness of the information extraction process. To address this, future work should improve data consistency for DVC processes. The algorithm's flexibility in constraint management, vehicle-specific configuration capabilities, and rapid integration of new derivatives or EE-architectures position it for further development. Expanding these capabilities could lead to even more efficient manufacturing systems that adapt dynamically to changing production requirements. Currently, the implementation does not support real-time updates or dynamic rescheduling in response to disruptions. Additional integration of real data such as the actual duration of processes or feedback from employees from different assembly environments would highlight potential weaknesses of the algorithm and reveal further optimization opportunities. The ability to adapt the precedence graph of DVC in real time, ideally using feedback from a failed process, would represent a useful further development and integration.


\begin{thebibliography}{00}

\bibitem{01}
H. Askaripoor, M. Hashemi Farzaneh, and A. Knoll, "E/E Architecture Synthesis: Challenges and Technologies," Electronics, vol. 11, no. 4, p. 518, 2022, DOI: 10.3390/electronics11040518.

\bibitem{02}
W. Zimmermann and R. Schmidgall, Bussysteme in der Fahrzeugtechnik. Wiesbaden, Germany: Springer Vieweg, 2014, DOI: 10.1007/978-3-658-02419-2.

\bibitem{03}
R. T. Lutchen, "Methode zur Identifizierung der Fahrzeug-Netzwerk-Architektur," in Optimierung der Fahrzeugdiagnose durch eine cloudbasierte Methode zur Identifikation der Datennetze mit künstlicher Intelligenz, Wiesbaden, Germany: Springer Fachmedien Wiesbaden, 2023, pp. 55-112, DOI: 10.1007/978-3-658-43113-6\_4.

\bibitem{patent1}
"Verfahren zur Programmierung eines Steuergeräts eines Kraftfahrzeugs," DE Patent DE102015203776A1, 2015.

\bibitem{patent2}
"Verfahren und Vorrichtung zur Programmierung eines Steuergeräts eines Fahrzeugs, insbesondere eines Kraftfahrzeugs," DE Patent DE10153447A1, 2001.

\bibitem{patent3}
"Verfahren und Vorrichtung zur fahrzeugtypischen Programmierung von Fahrzeugsteuergeräten," DE Patent DE10107263A1, 2001.

\bibitem{patent4}
"Inbetriebnahme-Steuergerät eines Verbunds aus Steuergeräten eines Kraftfahrzeugs und Verfahren zur Inbetriebnahme von Steuergeräten," DE Patent DE10107263A1, 2017.

\bibitem{pred1}
H. Klindworth, C. Otto, and A. Scholl, "On a learning precedence graph concept for the automotive industry," Eur. J. Oper. Res., vol. 217, no. 2, pp. 259–269, Sep. 2011, DOI: 10.1016/j.ejor.2011.09.024.

\bibitem{pred2}
C. Otto and A. Otto, "Multiple-source learning precedence graph concept for the automotive industry," Eur. J. Oper. Res., vol. 234, no. 1, pp. 253–265, Oct. 2013, DOI: 10.1016/j.ejor.2013.09.034.

\bibitem{pred3}
K. R. Antani, B. Pearce, L. Mears, R. Renu, M. E. Kurz, and J. Schulte, "Application of system learning to precedence graph generation for assembly line balancing," in Proc. ASME Int. Manuf. Sci. Eng. Conf., Jun. 2014, DOI: 10.1115/msec2014-3906.

\bibitem{pred4}
W. Kern, Methodik zur Gestaltung eines modularen Montagesystems, in Modulare Produktion, Wiesbaden, Germany: Springer Vieweg, 2021, pp. 149–222. doi: 10.1007/978-3-658-36300-0\_5.

\bibitem{sched1}
S. König et al., "Flexible scheduling of diagnostic tests in automotive manufacturing," Flex. Serv. Manuf. J., vol. 35, pp. 320–342, 2023, DOI: 10.1007/s10696-021-09438-3.

\bibitem{sched2}
R. Herberth, S. Körper, T. Stiesch, F. Gauterin, and O. Bringmann, "Automated Scheduling for Optimal Parallelization to Reduce the Duration of Vehicle Software Updates," IEEE Trans. Veh. Technol., vol. 68, no. 3, pp. 2921-2933, Mar. 2019, DOI: 10.1109/TVT.2019.2895109.

\bibitem{DIN72552}
Deutsches Institut für Normung e.V., "DIN 72552: Bezeichnungen für elektrische und elektronische Bauteile," Beuth Verlag, 2018.

\bibitem{ie1}
S. Singh, "Natural Language Processing for Information Extraction," arXiv:1807.02383, 2018.

\bibitem{ie2}
N. Arabadzhieva - Kalcheva and I. Kovachev, "Comparison of BERT and XLNet accuracy with classical methods and algorithms in text classification," 2021 International Conference on Biomedical Innovations and Applications (BIA), Varna, Bulgaria, 2022, pp. 74-76, doi: 10.1109/BIA52594.2022.9831281.

\bibitem{fuzzy1}
J. Ren et al., "Matching Algorithms: Fundamentals, Applications and Challenges," IEEE Trans. Emerg. Topics Comput. Intell., vol. 5, no. 3, pp. 332-350, Jun. 2021, DOI: 10.1109/TETCI.2021.3067655.

\bibitem{fuzzy2}
P. J. Rao, K. N. Rao, S. Gokuruboyina, and K. N. Neeraja, "An Efficient Methodology for Identifying the Similarity Between Languages with Levenshtein Distance," in Lect. Notes Electr. Eng., 2024, pp. 161–174, DOI: 10.1007/978-981-99-7137-4\_15.

\bibitem{bert1}
C. Sun, X. Qiu, Y. Xu, and X. Huang, "How to Fine-Tune BERT for Text Classification?," in Lect. Notes Comput. Sci., 2019, pp. 194–206, DOI: 10.1007/978-3-030-32381-3\_16.

\bibitem{bert2}
K. Taneja and J. Vashishtha, "Comparison of Transfer Learning and Traditional Machine Learning Approach for Text Classification," in Proc. 9th Int. Conf. Comput. Sustainable Global Develop. (INDIACom), Mar. 2022, pp. 195–200, DOI: 10.23919/indiacom54597.2022.9763279.

\bibitem{distilbert}
V. Sanh, L. Debut, J. Chaumond, and T. Wolf, "DistilBERT, a distilled version of BERT: smaller, faster, cheaper and lighter," arXiv:1910.01108, 2020.

\bibitem{gaussiannb1}
M. Kowsari, K. J. Meimandi, M. Heidarysafa, S. Mendu, L. Barnes, and D. Brown, "Text Classification Algorithms: A Survey," Information, vol. 10, no. 4, p. 150, 2019.

\bibitem{svm1}
Y. Zhang and X. Y. Lee, "A Comparison of Methods for Multi-Class Text Classification," J. Mach. Learn. Res., vol. 21, no. 76, pp. 1-22, 2020.

\bibitem{flexray}
R. Makowitz and C. Temple, "Flexray - A communication network for automotive control systems," in Proc. IEEE Int. Workshop Factory Commun. Syst., Turin, Italy, 2006, pp. 207-212, DOI: 10.1109/WFCS.2006.1704153.

\bibitem{lin}
Y. Xu, J. Wang, W. Chen, J. Tao, and Q. Liu, "Application of LIN Bus in Vehicle Network," in Proc. IEEE Int. Conf. Veh. Electron. Safety, Shanghai, China, 2006, pp. 119-123, DOI: 10.1109/ICVES.2006.371566.

\bibitem{most}
E. Zeeb, "Optical data bus systems in cars: current status and future challenges," in Proc. 27th Eur. Conf. Opt. Commun., Amsterdam, Netherlands, 2001, pp. 70-71, DOI: 10.1109/ECOC.2001.989436.

\bibitem{milp}
I. Dimény and T. Koltai, "Comparison of MILP and CP models for balancing partially automated assembly lines," Cent. Eur. J. Oper. Res., vol. 32, pp. 945–959, 2024, DOI: 10.1007/s10100-023-00885-x.

\end{thebibliography}
\end{document}